\pdfoutput=1

\documentclass[11pt]{article}

\usepackage[final]{acl}

\usepackage{times}
\usepackage{latexsym}
\usepackage{amsmath}
\usepackage{booktabs}       
\usepackage{multirow}
\usepackage{makecell}
\usepackage{graphicx}
\usepackage{xcolor}
\usepackage{comment}
\usepackage{hyperref}

\usepackage[T1]{fontenc}

\usepackage[utf8]{inputenc}

\usepackage{microtype}

\usepackage{inconsolata}
\newcommand{\ours}{\texttt{RE-RAG}}
\newcommand{\ourmodule}{Relevance estimator}
\newcommand{\sourmodule}{RE}
\newcommand{\sref}[1]{\S\ref{#1}} %

\definecolor{RoseQuartzBg}{HTML}{F7CAC9}
\definecolor{RoseQuartz}{HTML}{F5A798}
\definecolor{Serenity}{HTML}{92A8D1}
\definecolor{OrangeRed}{rgb}{1.0, 0.27, 0.0}
\definecolor{Red}{rgb}{1.0, 0.0, 0.0}
\definecolor{Turquoise}{HTML}{0F4C81}

\NewDocumentCommand{\ks}{ mO{} }{\textcolor{blue}{\textsuperscript{\textit{KS}}\textsf{\textbf{\small[#1]}}}}
\NewDocumentCommand{\jy}{ mO{} }{\textcolor{RoseQuartz}
{\textsuperscript{\textit{jy}}\textsf{\textbf{\small[#1]}}}}
\NewDocumentCommand{\jyc}{ mO{} }{\textcolor{Turquoise}
{\textsuperscript{\textit{jcomment}}\textsf{\textbf{\small[#1]}}}}
\NewDocumentCommand{\ksc}{ mO{} }{\textcolor{OrangeRed}
{\textsuperscript{\textit{KScom}}\textsf{\textbf{\small[#1]}}}}

\title{RE-RAG: Improving Open-Domain QA Performance and Interpretability with Relevance Estimator in Retrieval-Augmented Generation}

\author{Kiseung Kim \hspace{0.8cm} Jay-Yoon Lee\thanks{Corresponding author} \\
  Graduate School of Data Science, Seoul National University \\
  \texttt{\{kkskp, lee.jayyoon\}@snu.ac.kr} \\}

\begin{document}

\maketitle
\begin{abstract}
The Retrieval Augmented Generation (RAG) framework utilizes a combination of parametric knowledge and external knowledge to demonstrate state-of-the-art performance on open-domain question answering tasks. However, the RAG framework suffers from performance degradation when the query is accompanied by irrelevant contexts. In this work, we propose the RE-RAG framework, which introduces a relevance estimator (RE) that not only provides relative relevance between contexts as previous rerankers did, but also provides confidence, which can be used to classify whether given context is useful for answering the given question. We propose a weakly supervised method for training the RE simply utilizing question-answer data without any labels for correct contexts. We show that RE trained with a small generator (sLM) can not only improve the sLM fine-tuned together with RE but also improve previously unreferenced large language models (LLMs). Furthermore, we investigate new decoding strategies that utilize the proposed confidence measured by RE such as choosing to let the user know that it is “unanswerable” to answer the question given the retrieved contexts or choosing to rely on LLM’s parametric knowledge rather than unrelated contexts. \footnote{Code is available at \href{https://github.com/kkskp/re-rag}{here}}
\end{abstract}

\section{Introduction}

In recent years, the retrieval augmented generation framework has shown promising progress in natural language generation, specifically on knowledge-intensive tasks. This approach has been studied in many forms, from traditional RAG \cite{lewis2020retrieval}, which aggregates answers from multiple contexts using document relevance scores as weights, to approaches like RALM \cite{ram2023context}, which simply utilizes concatenated context as an in-context learning approach for large-language models (LLMs). Retrieval augmented generation enhances the model’s faithfulness and reliability by leveraging nonparametric knowledge on top of parametric knowledge \cite{luo2023augmented}. In particular, the RAG framework has the advantage of being easily adaptable to modern LLMs \cite{brown2020language, touvron2023llama}. These advantages have sparked a significant amount of new research \cite{asai2023self, lin2023ra, shi2023replug} focused on the RAG framework.

Despite the great potential of the retrieval augmented generation framework, if the language model is provided with contexts that are not relevant to the query, it will be distracted by these inappropriate contexts, negatively affecting the accuracy of the answers \cite{yoran2023making}. While retrievers or re-rankers in existing research have been effective at measuring the relative ranking across contexts to a query, these modules often fail to determine whether top-ranked contexts are actually relevant to the query or not. Furthermore, if a precise relevance score is not used in the traditional RAG framework, it can cause problems such as directing attention to documents that are less likely to answer the query.



In this work, we propose the \texttt{RE-RAG} framework, which extends traditional RAG by incorporating a relevance estimator (\texttt{RE}) to simultaneously measure the precise relative relevance between retrieved contexts and evaluate their confidence, which can be used to classify whether given context is useful for answering the given question. By more accurately measuring the relative relevance between contexts, RE computes precise relevance scores suitable for weighted aggregated answers in the traditional RAG framework and also acts as an efficient reranker. RE trained on a small generator (sLM) not only benefits sLM fine-tuned together with RE but can also be separated and applied to LLMs as well, benefiting both.


By explicitly classifying whether the context is useful for answering the query, the confidence of context measured by RE provides various decoding strategies. If the retrieved context set is irrelevant, we can choose to classify the query as ``unanswerable'', while maintaining most of the accuracy for the answerable set. Additionally, if a low-confidence context set is retrieved, which will likely result in wrong answers by parroting the context as is \cite{jia2017adversarial}, we can instead selectively leverage the LLM’s parametric knowledge to improve answer accuracy in most cases.


The main contributions of our work are: 
\begin{enumerate}

\item We propose a new framework called \textbf{\ours} by adding an external  Relevance Estimator (\texttt{RE}) module. We further suggest a weak supervision training method that can train \texttt{RE} without explicit labeled data on question-context compatibility. (\sref{subsec:method})



\item We demonstrate that \texttt{RE-RAG}, enhanced with
\texttt{RE}, significantly improves upon the existing
RAG. Additionally, we show that \texttt{RE} trained on a small language model can improve the answer performance of LLMs. (\sref{subsec:main_result})

\item We propose to use the confidence level of the context set measured by \texttt{RE} to answer ``unanswerable'' for unanswerable context sets with minimal negative effects, or to complement LLM's parametric knowledge. (\sref{subsec:decode_strategy})
\end{enumerate}
\section{Method}
\label{sec:our_method}
In this section, after reviewing the traditional RAG framework, we present the RE-RAG model combined with our relevance estimator.

\subsection{Traditional RAG overview}
\textbf{Retriever} Retriever searches for information in an external knowledge base and returns a related context set ${\mathbf C}_{i}$. In general, RAG systems use a bi-encoder type retriever such as DPR \citep{karpukhin-etal-2020-dense}, which is effective and fast in retrieving information. A question ${\mathbf q}_i$ $\in$ ${\mathbf Q}$ and a context ${\mathbf c}_j$ $\in$ ${\mathbf C}_{i}$ are input to the encoder independently to obtain an embedding of $\mathbf{Emb}_q = Encoder(q_i)$, $\mathbf{Emb}_c = Encoder(c_j)$. The similarity score ${\mathbf S}_{i,j} = \mathbf{Emb}_q \cdot \mathbf{Emb}_c$ is calculated from the obtained embedding and then used to perform top-$k$ context retrieval.

\noindent\textbf{Generator} Generators that utilize the sequence-to-sequence model typically take a question and context as input and produce an answer $\mathbf{y}_{i,j}$ with probability ${\mathbf P}_{G}({\mathbf y}_{i,j}|{\mathbf q}_i, {\mathbf c}_j)$.

\noindent\textbf{Answer marginalization} RAG \citep{lewis2020retrieval} introduced the answer generation models of RAG-sequence and RAG-token. We focus on the RAG-sequence model which marginalizes probability of 
$y_l\in\mathcal{Y}_i$ where $\mathcal{Y}_i$ is an aggregated set of $\mathbf{y}_{i,j}$.
which achieves higher performance than the RAG-token model and ensures the interpretability of the answer generation process. 
Individually generated answers $\mathbf{y}_{i,j}$ per $\mathbf{c}_j$ are marginalized as $\mathbf{y}_l$ using the similarity score ${\mathbf S}_{i,j}$ as shown in eq.\eqref{eq:RAG-seq}.

{\small
\begin{gather}    
\label{eq:RAG-score}
\mathbf{P}_{R}(\mathbf{S}_{i,j}) = \frac{e^{\mathbf{S}_{i,j}}}{\sum_{k}e^{\mathbf{S}_{i,k}}} \\[6pt]
\label{eq:RAG-seq}
\mathbf{P}_{a}(\mathbf{y}_l|\mathbf{q}_i, \mathbf{C}_i) = \sum_{j}  \mathbf{P}_{R}({\mathbf S}_{i,j}) \cdot {\mathbf P}_{G}({\mathbf y}_{l}|{\mathbf q}_i, {\mathbf c}_j)  
\end{gather}
}
\begin{figure}[t]
  \centering
  \includegraphics[width=0.45\textwidth]{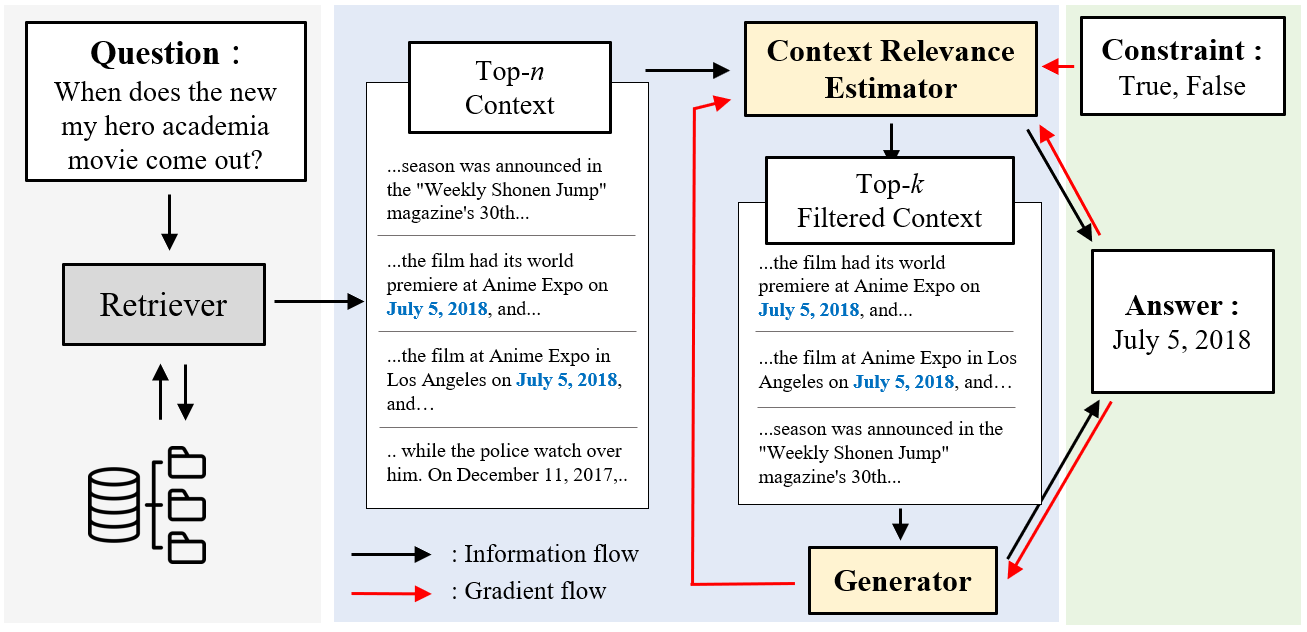}
  \caption{Overview of our proposed RE-RAG framework. The black lines represent the flow of information and the red lines represent the flow of gradients.}
  \label{fig:picture_1}
\end{figure}
\subsection{RE-RAG framework}
\label{subsec:method}
The retriever similarity score $\mathbf{S}_{i,j}$ is trained to achieve high recall when retrieving multiple contexts, however, it was not initially designed to provide fine-grained relevancy score $\mathbf{P}_{R}(\mathbf{S}_{i,j})$ for aiding RAG generation steps in eq.\eqref{eq:RAG-seq}.
To address this issue, we propose a relevance estimator (RE) that re-ranks contexts and provides precise relevance scores to the generator.

\paragraph{Relevance Estimator}
Relevance estimator (RE) measures the relevance between a question and context. 
We utilize a similar architecture to \citet{nogueira2020document}
which utilizes a sequence-to-sequence model as a passage reranker. 

Our \sourmodule\ receives the same input of question and context as the generator, but is trained to generate a \textbf{classification token} ("true" or "false") based on the relevance of the context to the input question. We normalize the probability of generating "true" and "false" tokens to get the final probability of generating the classification token. The obtained probability of a "true" token can independently be an indicator of the relevance of a single context to a given question. When comparing between multiple contexts, the "true" token probability can be converted to logit and used as the relevance score of the retrieved context.

{\small
\begin{align}
\label{eq:REscore}
\mathbf{RE}_{i,j} = \frac{\mathbf{P}(``true"|\mathbf{q}_i, \mathbf{c}_j)}{\mathbf{P}(``true"|\mathbf{q}_i, \mathbf{c}_j) + \mathbf{P}(``false"|\mathbf{q}_i, \mathbf{c}_j)}
\end{align}
}
\paragraph{Reranking of contexts by relevance} 
With the trained relevance estimator  $\texttt{RE}$,
we can rerank contexts in the initial retrieved set $\mathbf{C}_{i}$ by their relevance and only take top-$k$ contexts to redefine $\mathbf{C}_{i}$ before the answer-generation step. With a precise relevance score from $\texttt{RE}$, we can expect the \texttt{RE-RAG} to be more efficient, i.e. stronger performance with lower computation (see \sref{subsec:ablation_num_context}). 

\paragraph{Answer marginalization with context RE}
The question and context are concatenated and input to the generator model, and the generator generates ${\mathbf P}_{G}({\mathbf y}_{i,j}|{\mathbf q}_i, {\mathbf c}_j)$ per question. We replace the probability distribution $\mathbf{P}_{R}({\mathbf S}_{i,j})$ in eq.\eqref{eq:RAG-seq} with the relevance scores from context RE to form eq.\eqref{eq:RE-RAG-seq} as following:

{\small
\begin{gather}
\sigma(\mathbf{RE}_{i,j}) = \log \left( \frac{\mathbf{RE}_{i,j}}{1-\mathbf{RE}_{i,j}} \right)\tag{4} \\[6pt]
\mathbf{P_{RE}}(\mathbf{q}_i, \mathbf{c}_{j}) 
 = \frac{e^{\sigma(\mathbf{RE}_{i,j})}}{\sum_{k}e^{\sigma(\mathbf{RE}_{i,k})}} \tag{5} \\[6pt]
\mathbf{P}_{a}(\mathbf{Y}_i|\mathbf{q}_i, \mathbf{C}_i) = \sum_{j} 
\mathbf{P_{RE}}(\mathbf{q}_i, \mathbf{c}_{j}) 
\cdot \mathbf{P}_{G}(\mathbf{y}_{i,j}|\mathbf{q}_i, \mathbf{c}_j). \label{eq:RE-RAG-seq} \tag{6}
\end{gather}
}
We can expect higher performance 
with the marginalized answer $\mathbf{y}_l$ if RE can provide an accurate relevance distribution $\mathbf{P_{RE}}$ (see \sref{subsec:ablation}). 

\subsection{Joint training of RE-RAG}
\label{subsec:joint_train}
We propose to utilize three different types of losses to train \ours\ with our proposed relevance estimator. First, to train the generator model, we use a loss that combines the commonly used negative likelihood loss for ground truth $\mathbf{a}_i$ with a probability that represents the relevance of the question and context.

{\small
\begin{align}
\mathbf{L}_{\text{gen}} = -\sum_{i,j} \log \left( \mathbf{P_{RE}}(\mathbf{q}_i, \mathbf{c}_{j}) \cdot \mathbf{P}_{G}(\mathbf{a}_{i}|\mathbf{q}_i, \mathbf{c}_j) \right) \tag{7}
\end{align}
}

$\mathbf{L}_{\text{gen}}$ simultaneously adjusts the probability of generating the classification token for the relevance estimator while training the generator.

Second, to obtain a learning signal for training the relevance estimator, we calculate the log-likelihood loss of the generator per retrieved context and compute its distribution across contexts as follows:

\begin{gather}
\mathbf{F}_{i,j} = \log(\mathbf{P}_{G}(\mathbf{a}_{i} |\mathbf{q}_{i}, \mathbf{c}_{j})) \tag{8} \\
\mathbf{Q}_{G}(\mathbf{q}_i, \mathbf{c}_{j}) = \frac{e^{\mathbf{F}_{i,j}}}{\sum_{k}e^{\mathbf{F}_{i,k}}}. \tag{9}
\end{gather}

The log-likelihood loss varies depending on whether an answer can be inferred from the input context. Therefore, applying the softmax function to the log-likelihood loss values yields a probability distribution that represents the relevance between the given set of contexts and the question. We do not leverage any labeled data that entails the relevance of questions and contexts.


$\mathbf{Q}_{G}(\mathbf{q}_i, \mathbf{c}_{j})$ represents relative relevance between $\mathbf{q}_i$ and $\mathbf{c}_j$

We calculate the KL-divergence loss between the probability distributions of the generator and the \sourmodule, and use this loss to train the model.

\begin{gather}
\mathbf{L}_{\text{re}} = D_{\text{KL}}(\mathbf{P}_{RE}(\mathbf{q}_i, \mathbf{c}_{j}) || \mathbf{Q}_{G}(\mathbf{q}_i, \mathbf{c}_{j})) \tag{10}
\end{gather}

Lastly, in addition to applying a training loss on the probability of generating the classification token, we need to set an additional loss to prevent the \sourmodule\ from generating tokens other than the classification token. To do this, we utilize the additional loss as the sum of the  probability of \sourmodule\ of generating all tokens other than classification token.

\begin{gather}
\mathbf{L}_{\text{tok}} = \sum_{t \in T \setminus \{\text{"true"}, \text{"false"}\}} \mathbf{P}(t|\mathbf{q}_i, \mathbf{c}_k) \tag{11}
\end{gather}

To train an effective system, the two models are trained jointly utilizing all three losses as follows:
\begin{gather}
\mathbf{L}_{\text{tot}} = \mathbf{L}_{\text{gen}}+ \mathbf{\alpha}_1 \mathbf{L}_{\text{re}}+ \mathbf{\alpha}_2 \mathbf{L}_{\text{tok}} \tag{12}
\end{gather}
where $\mathbf{\alpha}_1$ and $\mathbf{\alpha}_2$ are hyperparameters that act as scaling factors to balance the impact of each loss.

\begin{table*}[t]
\centering
\scalebox{0.6}{
\begin{tabular}{lccccccccc}
\toprule
\multirow{2}{*}{\textbf{Model}} & \multirow{2}{*}{\textbf{Extra}} & \multirow{2}{*}{\textbf{Generator}} & \multicolumn{3}{c}{\textbf{NQ}} & \multicolumn{3}{c}{\textbf{TQA}} & \multirow{2}{*}{\textbf{\# Contexts}} \\
\cmidrule(lr){4-6} \cmidrule(lr){7-9}
 & & & \textbf{EM} & \textbf{Acc} & \textbf{F1} & \textbf{EM} & \textbf{Acc} & \textbf{F1} &  \\
\midrule
\multicolumn{10}{c}{\textit{Small language models ($\leq 2B$)}} \\
\midrule
RAG \citep{lewis2020retrieval}            & - & 445M & 44.5 & -  & - & 56.8  & -  & - & 50 \\
\hline
$\text{FiD}_{base}$ \citep{izacard2021leveraging}       & - & 220M & 48.2 & -  & -  & 65.0 & -  & -  & 100 \\
$\text{FiD}_{large}$ \citep{izacard2021leveraging}        & - & 770M & 51.4 & -  & -  & 67.6 & -  & -  & 100 \\ 
$\text{FiD-KD}_{base}$ \citep{izacard2021distilling}      & - & 220M & 50.1 & -  & -  & \underline{69.3} & -  & -  & 100 \\ 
$\text{FiD-KD}_{large}$ \citep{izacard2021distilling}    & - & 770M & \underline{54.4} & -  & -  & \textbf{72.5} & -  & -  & 100 \\ 
ReAtt \citep{jiang2022retrieval}           & - & 770M & \textbf{54.7} & -  & -  & -  & -  & - & 100\\ 
\hline
$\text{FiD-KD}_{base}$ \citep{izacard2021distilling}      & - & 220M & 48.6 & -  & -  & 67.4 & -  & -  & 25 \\ 
$\text{FiD-KD}_{large}$ \citep{izacard2021distilling}    & - & 770M & 53.9 & -  & -  & \underline{71.2} & -  & -  & 25 \\ 
R2-D2 \citep{fajcik2021r2}            & 125M & 1.04B & \textbf{55.9} & -  & -  & 69.9 & -  & -  & 25 \\ 
$\text{\ours}_{base}$ & 220M & 220M & 49.9 & 53.1  & 56.9 & 68.2 & 70.0 & 73.6 & 25\\ 
$\text{\ours}_{Flan-base}$ & 220M & 220M & 51.9 & 55.2 & 58.9 & 70.1 & 72.0 & 75.8 & 25\\ 
$\text{\ours}_{large}$ & 770M & 770M & 54.0 & 56.7 & 61.0 & 70.2 & 71.7 & 75.9 & 25\\ 
$\text{\ours}_{Flan-large}$ & 770M & 770M & \underline{55.4} & 58.3 & 62.5 & \textbf{72.9} & 74.4 & 78.7 & 25\\ 

\midrule
\multicolumn{10}{c}{\textit{Large language models ($\geq 7B$)}} \\
\midrule

$\text{Self-RAG}_{7B}$ \citep{asai2023self}    & - & 7B & - & - & - & - & 66.4 & - & 5 \\
$\text{Self-RAG}_{13B}$ \citep{asai2023self}    & - & 13B & - & - & - & - &69.3 & - & 5 \\
$\text{Llama2}_{7b} + \texttt{RE}$  & 770M & 7B & 45.7 & 48.4 & 54.3 & 67.1 &  \underline{70.1} & 73.3 & 5 \\
$\text{Llama2}_{13b} + \texttt{RE}$  & 770M & 13B & 46.6 & 49.8 & 55.6 & 70.8 & \textbf{73.2} & 77.2 & 5 \\
\midrule 

$\text{RA-DIT}$ \citep{lin2023ra} & - & 65B & 43.9 & - & - & 75.1 & - & - & 10 \\
$\text{Llama3}_{8b} + \text{FiD-KD}_{ret}$  & - & 8B & 37.9(38.2) & 43.9(40.2) & 47.5(47.4) & 63.8(57.6) & 66.7(59.3) & 70.7(63.3) & 10 \\
$\text{Llama2}_{70b} + \text{FiD-KD}_{ret}$  & - & 70B & 38.1(40.7) & 43.0(47.4) & 47.5(50.8) & 63.5(66.3) & 66.4(71.4) & 70.0(73.2) & 10 \\
$\text{Llama3}_{70b} + \text{FiD-KD}_{ret}$  & - & 70B & 39.5(46.8) & 44.3(52.4) & 48.5(56.9) & 68.1(72.1) & 70.8(75.3) & 74.7(79.1) & 10 \\
$\text{ChatGPT} + \text{FiD-KD}_{ret} $     & - & 175B & 42.9(45.9) & 46.6(50.0) & 52.2(56.2) & 69.0(70.7) & 74.5(74.0) & 76.8(77.8) & 10 \\ 
$\text{Codex} + \text{REPLUG LSR}$ \citep{shi2023replug} & - & 175B & 45.5 & - & - & \textbf{77.3} & - & - & 10 \\
$\text{Llama3}_{8b} + \texttt{RE}$   & 770M & 8B & \underline{49.6} & 54.5 & 59.0 & 73.0 & \underline{75.4} & 79.3 & 10 \\
$\text{Llama2}_{70b} + \texttt{RE}$  & 770M & 70B & 48.0 & 52.0 & 57.6 & 72.4 & 74.8 & 78.6 & 10 \\
$\text{Llama3}_{70b} + \texttt{RE}$  & 770M & 70B & \textbf{50.8} & \underline{54.8} & \textbf{60.1} & \underline{75.5} & \textbf{77.7} & \textbf{81.7} & 10 \\
$\text{ChatGPT} + \texttt{RE} $     & 770M & 175B & 49.3 & \textbf{55.2} & \underline{59.6} & 72.6 & \textbf{77.7} & \underline{80.3} & 10 \\

\bottomrule
\end{tabular}}
\caption{EM scores on Natural Questions and TriviaQA datasets. The parameters of the generator and the extra module that evaluates a given context are listed separately. \# Contexts refer to the number of contexts utilized for inference. For an effective comparison, we divided the groups based on the size of the generator model and the number of contexts utilized for inference. Llama2 7B and 13B models were additionally tested with five contexts for a fair comparison with the Self-RAG \citep{asai2023self} baseline. Experiments on LLM ($\geq 7B$) followed the method of aggregating answers using relevance score weights per. However, in the case of applying the FiD-KD retriever to LLMs, we add one more number in the (right) to represent the zero-shot RALM method.  which concatenates contexts to generate answers. We provide this extra result in brackets to compare fairly with the FiD-KD retriever as its performance in the traditional RAG setting was incomparable due to its subpar performance. This shows that the FiD-KD score may be good for reranking but not a suitable relevance score for the traditional RAG method to perform well.
The bold is the best score in each group, and the underline is the second best. The bold and underline are only for figures that can be compared to the baseline.}
\label{tab:main_results}
\end{table*}

\section{Experimental Setup}
We evaluated the performance of our model on an open-domain QA dataset. In this section, we describe the dataset we used in our experiments and the details of our experiments.

\subsection{Dataset}
We evaluate our performance on two open-domain QA datasets:Natural Questions \citep{kwiatkowski-etal-2019-natural}, TriviaQA \citep{joshi2017triviaqa}. To train and evaluate our model, we utilize the context datasets retrieved for each question from NQ and TQA, as used in FiD-KD \citep{izacard2021distilling} and Akari \citep{asai2022evidentiality}. The dataset includes the top-20 training contexts, while the dev and test sets contain the top-100 contexts retrieved by the retriever. We used 20 contexts for training and the top-25 contexts extracted by the \sourmodule\ from the top-100 retrieved contexts for inference.

\noindent \textbf{Natural Questions} Natural Questions \citep{kwiatkowski-etal-2019-natural} is a dataset of real questions asked by users on the web. The dataset consists of questions collected from the web, a long answer that can be viewed as gold context for the question, and a short answer with a short span. The open-domain QA version dataset of Natural Questions is a dataset that collects only questions where the answer span of the short answer is 5 tokens or less in length. We use the NQ-open dataset.

\noindent \textbf{TriviaQA} TriviaQA \citep{joshi2017triviaqa} is a dataset of question-answer pairs collected from trivia enthusiasts. Each question and answer in the dataset has been reviewed by human annotators. We want to use the unfiltered version of TriviaQA dataset.

\subsection{Evaluation Metric}
The predicted answers are evaluated using \textbf{EM score}, a commonly used metric as in \citet{izacard2021leveraging}, \citet{rajpurkar2016squad}. The generated answers are normalized (e.g., lowercase, punctuation, article stripping) and compared to the correct answers in the dataset. We consider a generated answer to be correct if it exactly matches one of the correct answers in the given dataset after normalization.


We also provide F1 score and accuracy (Acc) as an additional evaluation metric as some previous paper only report Acc \citep{asai2023self}, which assesses whether the generated string contains the gold answer. These scores show a similar trend with the EM score, that RE-RAG outperforms the baseline methods. Nonetheless, since most baselines report EM scores exclusively,  our comparison is focused on EM scores.

\subsection{Baseline}
We investigate whether the performance of \ours\ is competitive with that of the FiD \citep{izacard2021leveraging}-based system. FiD has achieved excellent performance on the Question-Answering task, and the FiD-based application system also outperforms the RAG \citep{lewis2020retrieval}-based system on the QA task. 

We consider an additional baseline to compare the performance of \texttt{RE} when applied to LLMs. We compare the performance of \texttt{RE} and FiD-KD retriever when applied to LLMs. When applying the FiD-KD retriever to LLMs, we compared two methods: traditional RAG, which uses the retriever similarity score to perform answer marginalization, and RALM, which concatenates all context. 
When generating answers for individual contexts using the traditional RAG method, we used 8-shot examples, while the RALM method employed a zero-shot approach due to context length limitations. Furthermore, we compare our performance with other studies \cite{asai2023self, lin2023ra, shi2023replug} that have implemented RAG in LLMs.

\subsection{Model}
The two components of our framework, \sourmodule\ and the generator, utilize the T5 model \citep{raffel2020exploring} and Flan-T5 \citep{chung2024scaling}. We utilize the base and large size models, and explore three different model sizes depending on the combination of the two models.

Additionally, we utilize Llama2 (7B, 13B, 70B), Llama3 \footnote{\href{https://github.com/meta-llama/llama3}{https://github.com/meta-llama/llama3}} (8B, 70B), and ChatGPT (``gpt-3.5-turbo-0125'' version) as generators to assess if RE brings performance improvements when applied to LLMs. In our experiments, the LLMs used as generators are not fine-tuned for the downstream task.

\section{Experiment Results}
We investigate the QA performance of the RAG system with our newly proposed relevance estimator (RE). In addition to the QA performance of the whole system, we also examine the performance of the \sourmodule\ independently.

\subsection{Main Results}
\label{subsec:main_result}
The overall accuracy of our system on the two datasets (NQ and TQA) is shown in Table \ref{tab:main_results}. Compared to the traditional RAG, our system, \ours, performs better despite having the same total number of parameters. Our proposed RE improves the reliability of the RAG system by more accurately measuring the relevance between question and context. Our model performed competitively with models based on FiD structures\citep{izacard2021distilling, jiang2022retrieval, fajcik2021r2}. We also found that our methodology was more efficient than the instructed tuned T5.


The accuracy of the RE module when applied to Large Language Models (LLMs) is shown at the bottom of Table \ref{tab:main_results}. We only included the RAG-based model in our comparison because the FiD-based model is not applicable to LLMs due to structural differences. The RE module outperforms the FiD-KD retriever when applied to LLMs. When the RE module is applied to Llama2, it surpasses the Self-RAG, where the LMs themselves inspect the retrieved context and generated answers. In TQA, REPLUG with Codex scores slightly higher. The performance of TQA seems to depend more on the generator model than NQ (see Figure \ref{fig:confidence_score_graph} for a related discussion), and we believe that this is the reason for the performance difference with Codex. Our model performs better on NQ, which is a more knowledge intensive task.



\begin{table}[t]
\centering
\scalebox{0.70}{
\begin{tabular}{clcccc}
\toprule
\multirow{2}{*}{\textbf{Dataset}} & \multirow{2}{*}{\textbf{Model}} & \multicolumn{4}{c}{\textbf{Recall@k}} \\
& & R@1 & R@5 & R@10 & R@20 \\
\midrule
\multirow{4}{*}{NQ} 
 & FiD-KD & 49.4 & 73.8 & 79.6 & 84.3  \\
 & $\text{MonoT5}_{large}$ & 46.2 & 72.4 & 80.1 & 84.7  \\
 & $\text{\ours}_{base}$ & 59.5 & 77.8 & 82.7 & 85.5  \\
 & $\text{\ours}_{large}$ & \textbf{61.9} & \textbf{79.4} & \textbf{83.6} & \textbf{86.4}  \\
\midrule
\multirow{4}{*}{TQA} 
 & FiD-KD & 60.1 & 77.0 & 80.9 & 83.6  \\
 & $\text{MonoT5}_{large}$ & 64.7 & 79.7 & 82.9 & 84.8  \\
 & $\text{\ours}_{base}$ & 67.0 & 81.5 & 83.6 & 85.4  \\
 & $\text{\ours}_{large}$ & \textbf{70.4} & \textbf{82.2} & \textbf{84.4} & \textbf{86.1}  \\
\bottomrule
\end{tabular}}
\caption{Performance of RE as a re-ranker. The re-ranking performance for the top-100 contexts retrieved by the FiD-KD retriever is denoted by recall@k.}
\label{tab:re_reranker}
\end{table}

\begin{table}[t]
\centering
\scalebox{0.72}{
\begin{tabular}{clccc}
\toprule
\textbf{Dataset} & \textbf{Model} & Recall & Precision & F1 \\
\midrule
\multirow{4}{*}{NQ}
& FiD-KD & 73.2 & 21.9 & 33.7 \\
& $\text{MonoT5}_{large}$ & 10.3 & 31.0 & 15.5 \\
& $\text{\ours}_{base}$ & 51.3 & 33.9 & 40.9 \\
& $\text{\ours}_{large}$ & 45.9 & 38.3 & 41.7 \\
\midrule
\multirow{4}{*}{TQA}
& FiD-KD & 64.3 & 24.5 & 35.5\\
& $\text{MonoT5}_{large}$ & 27.2 & 34.2 & 30.3 \\
& $\text{\ours}_{base}$ & 38.9 & 46.7 & 42.5 \\
& $\text{\ours}_{large}$ & 39.0 & 43.2 & 41.0 \\

\bottomrule
\end{tabular}}
\caption{Classification results for context sets that do not contain an answer within the top-25 context set. We used cosine similarity for FiD-KD's retriever and ``true'' token probability for our method and MonoT5.}
\label{tab:re_classification}
\end{table}

\subsection{Performance of RE as a reranker and unanswerable set classifier}
\label{subsec:ablation_num_context}
Table \ref{tab:re_reranker} shows the performance of our proposed \ours's \sourmodule\ as a reranker. For the Recall@k metric, we use the retrieval accuracy used by DPR \citep{karpukhin-etal-2020-dense}, FiD-KD \citep{izacard2021distilling}, and ColbertQA \citep{khattab-etal-2021-relevance}. Although the comparison retriever has been enhanced through knowledge distillation methods using FiD attention scores, our proposed \sourmodule\ still demonstrated superior performance. In particular, RE performs better as the number of contexts decreases, which means that RE is more efficient when there are fewer contexts to utilize.

Table \ref{tab:re_classification} shows the performance of the context relevance estimator (RE) as a ``unanswerable'' set classifier. ``unanswerable'' set means that the context set of the top-25 contexts does not contain a gold answer in any context. For classification, we used the cosine similarity score of the hidden representation of the question and context for retriever and the probability of generating a ``true'' token by the model for RE and MonoT5 \citep{nogueira2020document}. For the optimal threshold, we searched for the value that maximizes F1 score in steps of 0.1 from 0.5 to 0.9 at dev set.

Our \sourmodule\ showed better ``unanswerable'' set classification performance than FiD-KD retriever or MonoT5 based on F1 score. Looking at the detailed performance, we found that the retriever performed better for recall, but the \sourmodule\ performed better for precision. This is because the retriever classified a large number of context sets as all ``unanswerable'' sets, while our proposed \sourmodule\ showed a good balance between precision and recall.

\begin{table}
\centering
\scalebox{0.70}{
\begin{tabular}{lllcc}
\toprule
\multirow{2}{*}{\textbf{Dataset}} & \multirow{2}{*}{\textbf{Model}} & \multirow{2}{*}{\textbf{Score}} & \multicolumn{2}{c}{\textbf{Answerable context set}}\\
& & & \textbf{O} & \textbf{X} \\
\hline
\multirow{4}{*}{NQ}
& $\text{\ours}_{base}$ & $\text{FiD-KD}$  &  58.3 $\rightarrow$ 32.7 & 73.4 \\
& $\text{\ours}_{base}$ & $\text{RE}$  &  58.3 $\rightarrow$ 54.9 & 51.3 \\
& $\text{\ours}_{large}$ & $\text{FiD-KD}$  & 61.5 $\rightarrow$ 34.9 & 71.3 \\
& $\text{\ours}_{large}$ & $\text{RE}$  & 61.5 $\rightarrow$ 57.9 & 45.9 \\
\midrule
\multirow{4}{*}{TQA}
& $\text{\ours}_{base}$ & $\text{FiD-KD}$  & 78.7 $\rightarrow$ 51.2 & 63.5 \\
& $\text{\ours}_{base}$ & $\text{RE}$  & 78.7 $\rightarrow$ 77.0 & 38.9 \\
& $\text{\ours}_{large}$ & $\text{FiD-KD}$  & 80.4 $\rightarrow$ 51.6 & 62.7  \\
& $\text{\ours}_{large}$ & $\text{RE}$  & 80.4 $\rightarrow$ 77.9 & 39.0  \\
\bottomrule
\end{tabular}
}
\caption{
We examine whether RE can successfully identify unanswerable scenarios where retrieved contexts do not hold true answers. 
\textbf{O} refers to the retrieval context set that contains true answers and \textbf{X} refers to the set without which we dim as \textit{unanswerable}.
Under the \textbf{X}, we denote the classification accuracy for the unanswerable set. Under the \textbf{O}, we denote the accuracy change as the RE thresholding will inevitably classify the context sets with answers as unanswerable.  Left of the arrow denotes original accuracy on \textbf{O} and the right denotes accuracy after RE score thresholding.}
\label{tab:re_unanswerable}
\end{table}

\section{Analysis}
\subsection{Exploring decoding strategies in low confidence context sets}
\label{subsec:decode_strategy}
In this section, we review two strategies that can be used when a context set with a low confidence score is retrieved. The confidence score for a context set is determined using the maximum value of the ``true'' token probability computed by RE for the contexts within the set. We examine the strategy of answering ``unanswerable'' when a low confidence context set is returned in a small Language Model (sLM), where parametric knowledge is scarce. Additionally, we examine the strategy of directly utilizing parametric knowledge in Large Language Models (LLMs), where parametric knowledge is abundant.

\textbf{Classify as “unanswerable”}
Table \ref{tab:re_unanswerable} shows the change in accuracy after letting the model respond with ``unanswerable'' when the retrieved context set has low confidence. For the confidence threshold value that determines whether the model should respond with ``unanswerable'', we chose the value that optimizes the classification performance as determined in Table \ref{tab:re_classification}. We evaluate the accuracy by dividing the entire test set into answerable sets, which contain at least one gold answer in the context set, and unanswerable sets, which contain none.

Our RE model shows relatively minor accuracy loss on the answerable set when responding with “unanswerable” for context sets measured with low confidence, but gains significant ability on the unanswerable set. In contrast, the FiD-KD retriever loses a substantial amount of accuracy on the answerable set when it responds with “unanswerable” for low-confidence context sets, resulting in a larger negative effect compared to our model. If we want to preserve the answerable set accuracy of the FiD-KD retriever, its ability to classify ``unanswerable'' is significantly reduced compared to RE (see Appendix \ref{appendix:E}).

\begin{table}
\centering
\scalebox{0.62}{
\begin{tabular}{clcccc}
\toprule
\textbf{P-Generator} & \textbf{R-Generator} &  \textbf{NQ} & \textbf{TQA} \\
\hline
\multirow{3}{*}{\makecell{$\text{Llama2}_{70b}$ \\ {\footnotesize (NQ: 31.1/TQA: 64.3)}}}
& $\text{Llama2}_{7b}$    &  46.2 {\footnotesize $\rightarrow$} \colorbox{orange}{45.9(-0.3)} & 68.0 {\footnotesize $\rightarrow$} \colorbox{lime}{69.3(+1.3)} \\
& $\text{Llama2}_{13b}$   &  47.3 {\footnotesize $\rightarrow$} \colorbox{orange}{46.5(-0.8)} & 71.5 {\footnotesize $\rightarrow$} \colorbox{lime}{72.1(+0.6)} \\
& $\text{Llama2}_{70b}$   &  48.0 {\footnotesize $\rightarrow$} \colorbox{orange}{46.9(-1.1)} & 72.4 {\footnotesize $\rightarrow$} \colorbox{lime}{72.9(+0.5)} \\
\midrule
\multirow{2}{*}{\makecell{$\text{Llama3}_{70b}$ \\ {\footnotesize (NQ: 41.3/TQA: 75.1)}}}
& $\text{Llama3}_{8b}$    & 49.6 {\footnotesize $\rightarrow$} \colorbox{lime}{49.8(+0.2)} & 73.0 {\footnotesize $\rightarrow$} \colorbox{lime}{75.4(+2.4)} \\
& $\text{Llama3}_{70b}$   & 50.8 {\footnotesize $\rightarrow$} 50.8(-) & 75.5 {\footnotesize $\rightarrow$} \colorbox{lime}{76.7(+1.2)} \\
\midrule
\multirow{2}{*}{\makecell{$\text{ChatGPT}$ \\ {\footnotesize (NQ: 37.7/TQA: 72.0)}}}
& \multirow{2}{*}{$\text{ChatGPT}$} &  \multirow{2}{*}{49.3 {\footnotesize $\rightarrow$} 49.3(-)} & \multirow{2}{*}{72.6 {\footnotesize $\rightarrow$} \colorbox{lime}{73.6(+1.0)}} \\
& & & \\
\bottomrule
\end{tabular}
}
\caption{
Change in EM scores when utilizing the LLM's parametric knowledge for low-confidence context sets. P-Generator model, which relies solely on its parametric knowledge, has EM scores shown below its name. R-Generator refers to a model that utilizes RAG. For both datasets, the confidence score threshold for model selection is set to 0.7. See appendix \ref{appenidx:D} for results on FiD-KD retriever.}
\label{tab:selective_decoding}
\end{table}

\begin{figure}[!t]
  \centering
  \includegraphics[width=0.34\textwidth]{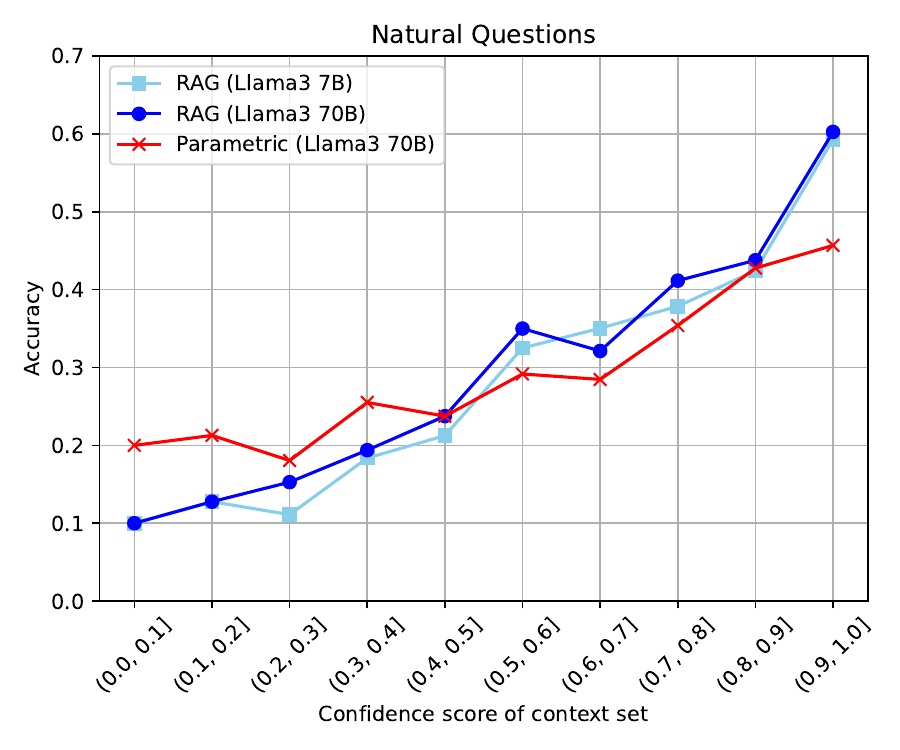}
  \includegraphics[width=0.34\textwidth]{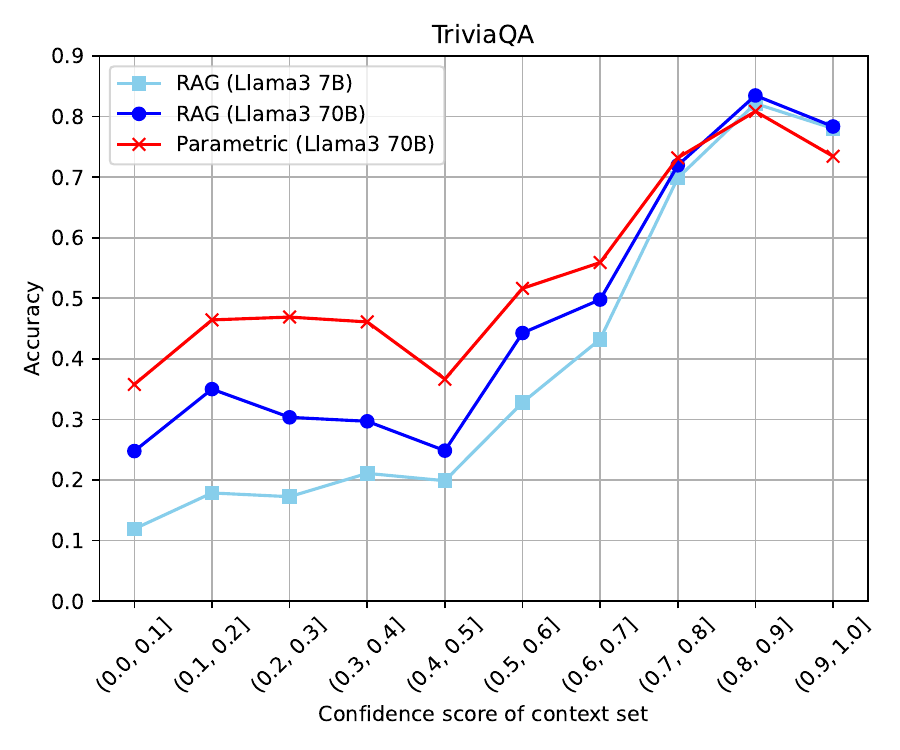}
  \caption{The relationship between confidence score and accuracy by model size. RAG means that the model utilizes contextual knowledge and Parametric means that the model utilizes only parametric knowledge without external knowledge.}
  \label{fig:confidence_score_graph}
\end{figure}    

\textbf{Selectively using parametric knowledge}
We further explore how we can effectively utilize the rich parametric knowledge of LLMs. When the confidence of the retrieved context is low, we examine a mixed strategy that optionally bypasses the context and relies solely on the parametric knowledge of the largest model to generate the correct answer. For the confidence threshold value that determines whether the model should answer using only parametric knowledge, we selected the value that optimizes classification performance as determined in Table \ref{tab:re_classification}. For each type of model, we utilize the one with the largest number of parameters as the parametric knowledge base.

Table \ref{tab:selective_decoding} shows the change in accuracy when decoding the answer using the mixed strategy. In most cases, our strategy achieves accuracy gains in TQA without significant losses in NQ, except in cases where parametric knowledge is particularly scarce, such as in NQ on Llama2. NQ is a more knowledge-intensive task compared to TQA, where there is less benefit from utilizing parametric knowledge.

When parametric knowledge can be used effectively, the mixed strategy achieves larger gains in smaller models, and the performance gap narrows compared to larger models. Figure \ref{fig:confidence_score_graph} illustrates the relationship between confidence score and accuracy by model size. At high confidence scores on the TQA dataset, small size models achieve similar accuracy to large size models. At low confidence scores, the difference in performance between small and large models becomes more pronounced. When using small size models, higher efficiency can be achieved by utilizing retrieval augmented generation only when a high confidence context set is retrieved, and selectively leveraging the parametric knowledge of large size models when a low confidence context set is retrieved.

\begin{table}
\centering
\scalebox{0.72}{
\begin{tabular}{lcc}
\toprule
\textbf{Model} & \textbf{NQ} & \textbf{TQA}\\
\hline
Baseline                       & 39.5 & 54.9 \\
Baseline w/ RE score           & 43.1 & 60.1 \\
Baseline w/ RE rerank         & 46.8 & 63.9 \\
Baseline w/ RE rerank, score  & 49.6 & 67.8 \\
\hline
$\text{RE-RAG}_{base}$         & 49.9 & 68.2 \\
\bottomrule
\end{tabular}}
\caption{An ablation study to decompose the effect of RE in RE-RAG. We compared the traditional RAG model without RE, with reranking of \sourmodule\ (RE rerank), with RE score in answer generation (RE score), and with both (RE rerank, score).}
\label{tab:ablation_table6}
\end{table}

\begin{table*}[t]
\centering
\scalebox{0.72}{
\begin{tabular}{clllll}
\toprule
\multirow{2}{*}{\textbf{Dataset}} & \multirow{2}{*}{\textbf{Model}} & \multicolumn{4}{c}{\textbf{Recall@k}} \\
& & R@1 & R@5 & R@10 & R@20 \\
\midrule
\multirow{4}{*}{NQ} 
 & FiD-KD\text{\scriptsize (NQ $\rightarrow$ NQ)} & 49.4 & 73.8 & 79.6 & 84.3  \\
 & FiD-KD\text{\scriptsize (TQA $\rightarrow$ NQ)}  & 35.9\text{\small $(-27.3\%)$} & 63.2\text{\small $(-14.4\%)$} & 73.1\text{\small $(-8.2\%)$} & 80.5\text{\small $(-4.5\%)$}  \\
 & $\text{\ours}_{large}$\text{\scriptsize (NQ $\rightarrow$ NQ)} & 61.9 & 79.4 & 83.6 & 86.4  \\
 & $\text{\ours}_{large}$\text{\scriptsize (TQA $\rightarrow$ NQ)} & 46.2\text{\small $(-25.4\%)$} & 71.6\text{\small $(-9.8\%)$} & 79.3\text{\small $(-5.1\%)$} & 83.9\text{\small $(-2.9\%)$}  \\
\midrule
\multirow{4}{*}{TQA} 
 & FiD-KD\text{\scriptsize (TQA $\rightarrow$ TQA)} & 60.1 & 77.0 & 80.9 & 83.6  \\
 & FiD-KD\text{\scriptsize (NQ $\rightarrow$ TQA)} & 47.6\text{\small $(-20.8\%)$} & 70.8\text{\small $(-8.1\%)$} & 76.8\text{\small $(-5.1\%)$} & 81.1\text{\small $(-3.0\%)$}  \\
 & $\text{\ours}_{large}$\text{\scriptsize (TQA $\rightarrow$ TQA)} & 70.4 & 82.2 & 84.4 & 86.1  \\
 & $\text{\ours}_{large}$\text{\scriptsize (NQ $\rightarrow$ TQA)} & 67.8\text{\small $(-3.7\%)$} & 80.2\text{\small $(-2.4\%)$} & 83.0\text{\small $(-1.7\%)$} & 85.1\text{\small $(-1.2\%)$}  \\
\bottomrule
\end{tabular}}
\caption{Change in rerank performance when applying the RE module and FiD-KD retriever to unseen datasets. The numbers in parentheses indicate the percentage drop on the unseen datasets.}
\label{tab:unseen_recall}
\end{table*}

\begin{table}
\centering
\scalebox{0.60}{  
\begin{tabular}{lccc}
\toprule
 \textbf{Model} & \textbf{NQ (EM/Acc)} & \textbf{TQA (EM/Acc)} & \textbf{\#Contexts} \\ 
\midrule
 $\text{Llama3}_{8b} +\text{FiD-KD}$ & 37.9/43.9 & 63.8/66.7 & 10 \\ 
 $\text{Llama3}_{8b} +\texttt{RE}$ & 49.6/54.5 & 73.0/75.4 & 10 \\ 
\midrule
 $\text{Llama3}_{8b} +\text{FiD-KD}_{\text{TQA}}$ & 30.3/34.7 & - & 10 \\ 
 $\text{Llama3}_{8b} +\texttt{RE}_{\text{TQA}}$ & 42.1/46.1 & - & 10 \\ 
\midrule
 $\text{Llama3}_{8b} +\text{FiD-KD}_{\text{NQ}}$ & - & 57.6/60.4 & 10 \\ 
 $\text{Llama3}_{8b} +\texttt{RE}_{\text{NQ}}$ & - & 70.3/73.0 & 10 \\ 
\bottomrule
\end{tabular}
}
\caption{Changes in answer performance when applying RE module and FiD-KD retriever to unseen datasets. In the model column, the subscript indicates the trained dataset, and NQ and TQA columns represent test data.}
\label{tab:unseen_accuarcy}
\end{table}

\subsection{Evaluation of relevance estimator on unseen dataset}
We evaluate the effectiveness of the relevance estimator (RE) module on unseen datasets that were not utilized during training from two perspectives: rerank performance and answer performance. The RE module and the baseline FiD-KD are trained only using a single dataset such as NQ (TQA). We analyze the changes in performance when applying the RE module and FiD-KD retriever to the new unseen dataset TQA (NQ).

Table \ref{tab:unseen_recall} compares the rerank performance of the RE module and the FiD-KD retriever on datasets that were not referenced in training. Overall, the RE module consistently shows a smaller performance drop compared to the FiD-KD retriever on these unseen datasets. In particular, when the RE module trained on Natural Questions is extended to the TriviaQA dataset (NQ$\rightarrow$TQA), both models show smaller performance drop than the opposite case (TQA$\rightarrow$NQ). However, the performance drop of the RE module is notably smaller (-3.7\%) than FiD-KD's (-20.8\%). This suggests that the RE module is more effective than FiD-KD retriever on unreferenced datasets when trained on datasets that are conducive to generalization.

Table \ref{tab:unseen_accuarcy} presents a comparison of the answer performance when the RE module and FiD-KD retriever are applied to the Llama3 8B generator on unreferenced datasets. The RE module consistently exhibits a smaller performance degradation compared to the FiD-KD retriever on both datasets, similar to the recall performance in Table \ref{tab:unseen_recall}.

\subsection{Ablation Study}
\label{subsec:ablation}
\textbf{Effectiveness of RE}
We perform an ablation study to investigate the effectiveness of the added \sourmodule\ in \ours. The effect of our proposed \sourmodule\ is twofold. First, it performs better re-ranking than the retriever, selecting more accurate context and passing it to the generator. Second, it calculates a more accurate relevance score than retriever's similarity score and uses it in the answer marginalization process. In Table \ref{tab:ablation_table6}, the performance of methods with each component of the \sourmodule\ added is presented, using a model that was trained with only the T5-base generator, after removing the \sourmodule, as the baseline.

We construct the following experiment to isolate the two effects. First, we apply the top 25 contexts from retriever and their similarity scores to the baseline model. Next, there are the top-25 contexts from the retriever with the \sourmodule's score applied (RE score) and the top-25 contexts from the \sourmodule\ with the retriever's similarity score applied (RE rerank). Finally, we compare the performance of applying the \sourmodule's top-25 contexts and score to the baseline model (RE rerank, score).

Both effects of the \sourmodule\ are found to be significant in improving the performance of the baseline model. This shows that not only the quality of the context input to the generator plays an important role, but also the score, which means the importance of each context. 

\textbf{Remove training components}
We investigate the impact of removing the regularization process in eq.(3) on the classification performance of RE while training on the $\texttt{RE-RAG}_{base}$ model. Table \ref{tab:true_prob_ab} shows how the “true” token probability level output by the RE changes when the normalization process is removed. It can be seen that when the normalization process is removed, RE can only perform the function of re-ranking but loses the function of measuring confidence. This is because the normalization process allows the model to adjust its output strictly between “true” and “false” tokens.

Table \ref{tab:L_re_ab} shows the difference in EM scores on the dev set when $L_{re}$ is removed from the training process. We observed that removing $L_{re}$ from the training process decreases answer performance. We believe that $L_{re}$ contributes to achieving more optimal performance by using loss information from generator to directly propagate the relative importance of contexts to the RE.

\begin{table}
\centering
\scalebox{0.85}{
\begin{tabular}{lcc}
\toprule
\textbf{Model} & \textbf{NQ} & \textbf{TQA}\\
\hline
Baseline                       & 0.435 & 0.561 \\
- normalization               & 0.0005 & 0.0002 \\
\bottomrule
\end{tabular}}
\caption{Average value of the probability that RE generates the "true" token for answerable contexts when the normalization process is removed.}
\label{tab:true_prob_ab}
\end{table}

\begin{table}
\centering
\scalebox{0.8}{
\begin{tabular}{lcc}
\toprule
\textbf{Model} & \textbf{NQ} & \textbf{TQA}\\
\hline
Baseline                       & 49.1 & 67.8 \\
- $L_{re}$                     & 48.0 & 66.7 \\
\bottomrule
\end{tabular}}
\caption{Difference in EM scores on the dev set when $L_{re}$ is removed from the training process.}
\label{tab:L_re_ab}
\end{table}

\section{Related Works}
Previous research has shown that the performance of Question Answering systems can be improved by utilizing external knowledge about questions \citep{chen-etal-2017-reading}. Methods for more accurate retrieval of external knowledge \cite{karpukhin-etal-2020-dense, khattab-etal-2021-relevance, gao-callan-2022-unsupervised} have been studied to make these systems more efficient. In open-domain QA, models that extract and use answers from retrieved documents have been studied \cite{karpukhin-etal-2020-dense, khattab-etal-2021-relevance, cheng-etal-2021-unitedqa}, but studies that utilize generative models such as T5 \citep{raffel2020exploring} or BART \citep{lewis-etal-2020-bart} have become more common \cite{lewis2020retrieval, izacard2021leveraging}. RAG and FiD achieved powerful performance in open-domain QA using different methods. Subsequently, models \cite{izacard2021distilling, fajcik2021r2, singh2021end, jiang2022retrieval} that leverage and improve upon the structural advantages of FiD have been proposed. For Atlas \citep{izacard2022few}, state-of-the-art performance was achieved through an improved retriever 
 \citep{izacard2021unsupervised} and scaling up the model. In the case of RAG, there is a study that improved performance by introducing a BERT \citep{devlin-etal-2019-bert}-based reranker \citep{glass-etal-2022-re2g}, but it utilized additional data and high-quality label data when training the reranker. 
 
Recently, large language models (LLMs) such as GPT \citep{brown2020language} and Llama \citep{touvron2023llama}, which have been developed in recent years, face limitations with FiD methods that require encoded data. Consequently, research on RAG models, which can directly input context, has received renewed attention. \cite{asai2023self, lin2023ra, shi2023replug} These approaches have achieved performance improvements by training a retriever, which can also be applied to LLMs, or by performing the review of questions and context within the model itself.

\section{Conclusion}
We propose the \ours\ framework, which extends traditional RAG by incorporating \texttt{RE} that can measure the relative relevance and confidence of contexts. We demonstrate that the \ours\ framework can enhance the performance of traditional RAG. We show that the RE module, as a detachable component, can be combined with modern large language models (LLMs) to improve their performance. Furthermore, we exploree some decoding strategies that leverage the confidence information measured by the RE module to either answer ``unanswerable'' or selectively utilize the parametric knowledge of the LLMs when a low confidence context set is retrieved. We hope that our research will inspire the exploration of various additional modules for retrieval-augmented generation.


\section{Limitation}
Our research has primarily focused on improving answer performance in single-hop QA tasks. We have not sufficiently verified the effectiveness of our proposed framework in multi-hop QA tasks. We believe that in the future, we can explore whether the RE-RAG framework can be extended to multi-hop QA.

In our work, we explored a decoding strategy that measures with confidence whether a context is truly useful for a query and classifies low confidence contexts as unanswerable. However, a truly unanswerable query is one where the query cannot be adequately answered even when utilizing the model's parametric knowledge. We believe that future research needs to be conducted to detect whether the parametric knowledge has knowledge that can adequately answer the query in order to finally classify the unanswerable problem.


\section{Acknowledgements}
This work was supported in part by the National Research Foundation of Korea (NRF) grant (RS-2023-00280883, RS-2023-00222663) and New Faculty Startup Fund from Seoul National University, and with the aid of computing resources from  Artificial Intelligence Industry Center Agency, and Google cloud platform research credits, and National Super computing Center with super computing resources including technical support (KSC-2023-CRE-0176).

\bibliography{custom}

\appendix
\section{Dataset Statistics}
Table \ref{tab:dataset_stats} shows the statistics for the Natural Questions and TriviaQA unfilitated datasets we used.
\begin{table}[!ht]
\centering
\scalebox{0.9}{
\begin{tabular}{lccc}
\hline
\textbf{Dataset} & \textbf{Train} & \textbf{Dev} & \textbf{Test}\\
\hline
Natural Questions & 79,168 & 8,757 & 3,610 \\ 
TriviaQA & 78,785 & 8,837 & 11,313 \\ 
\hline
\end{tabular}}
\caption{Dataset statistics for Natural Questions and TriviaQA}
\label{tab:dataset_stats}
\end{table}

\begin{table*}[t]
\centering\scriptsize
\begin{tabular}{p{0.18\textwidth}p{0.37\textwidth}p{0.25\textwidth}c}
\toprule
\textbf{Question} & \textbf{Context} & \textbf{Gold Answer} & \textbf{"True" prob} \\
\midrule
who played mark on the show the rifleman & ...\textbf{\textcolor{blue}{Mark McCain is the son of fictitious rancher Lucas McCain in the ABC Western television series "The Rifleman,"}} starring Chuck Connors, which ran from 1958 to 1963. Singer/actor and former Mouseketeer \textbf{\textcolor{blue}{Johnny Crawford was cast in the role}} and... & John Ernest Crawford &  0.987 \\
\midrule
when does the cannes film festival take place & ...2017 \textbf{\textcolor{blue}{Cannes Film Festival The 70th Cannes Film Festival took place from 17 to 28 May 2017}}, in Cannes, France ... & Cannes, France, usually in May &  0.994 \\
\midrule
how many strong verbs are there in german & ...\textbf{\textcolor{red}{Germanic strong verbs are commonly divided into 7 classes}}, based on the type of vowel alternation. This is in turn based mostly... & more than 200, more than 200 strong &  0.949 \\
\midrule
how many episodes of corrie has there been & ...The show airs six times a week: Monday, Wednesday and Friday 7:30-8 pm and 8:30-9 pm. Since 2017, \textbf{\textcolor{red}{ten sequential classic episodes}} of the series from 1986... & 9,436 &  0.147 \\
\bottomrule
\end{tabular}
\caption{The relevance measure of the question and context output by the \sourmodule. The first two show relevant contexts that contain the correct answer even if the context does not include exactly the same surface form compared to the true answer. The last two examples show irrelevant contexts that actually have high overlap with question tokens, however, without pertaining the correct answer.}
\label{tab:example_dataset}
\end{table*}

\section{Training Details}
We used T5-base with a parameter size of 223M and T5-large model with a parameter size of 770M as modulators in all experiments. We trained the $\text{\ours}_{base}$ system on 4 A6000 GPUs, while $\text{\ours}_{mixed}$ and $\text{\ours}_{large}$ were trained on 2 A100 and 4 A100 GPUs, respectively.

We used a constant learning rate of $10^-4$ for all sizes of RE-RAG systems. We used AdamW as the optimizer and weight decay was $10^-3$. For batch size, we used gradient accumulation for all sizes of models, resulting in an effective batch size of 64. For the hyperparameters that balance the proposed losses, we utilized the default value of 1 for both $\alpha_1$ and $\alpha_2$. We did not explore hyperparameters that achieve better performance due to time and limited computing resources.

For model selection, we evaluated every 1 epoch and selected the case with the highest answer accuracy of the dev set. The dev set answer accuracy was measured using the top-10 context of the \sourmodule. Since the answer accuracy of the top-10 context of the \sourmodule\ is similar to the answer accuracy of the top-25 context, this helped to save computational resources and time while still producing valid results.

\section{Effectiveness of the RE} 
We perform a qualitative analysis to see if our proposed relevance estimator (RE) is effectively classifying relevant contexts. Table 12 shows a few contexts in the NQ test set.

Some of the contexts that the \sourmodule\ predicts are highly relevant to the question even when they do not contain the exact ground truth answer. The first few examples in Table 3 are examples that are categorized as true context because they contain phrases that are semantically equivalent to the correct answer albeit not having the exact same form in the context. This shows that although the \sourmodule\ is trained to measure the relevance of a question to a context through a limited set of ground truth answers, it is actually capable of measuring a broader range of relevance. 

In addition to the examples above, there are cases where the \sourmodule\ misclassified contexts as containing the correct answer. As shown in the example in Table \ref{tab:example_dataset}, the \sourmodule\ classified the context containing ``the number of classes of strong verbs in German'' as the correct context for the question about ``the number of strong verbs in German'', which means that our \sourmodule\ is still limited in its ability to capture the fine-grained meaning of the question in the retrieved context. On the other hand, in the last example, for the question about ``the number of episodes'', it succeeded in classifying the context containing ``the number of classical episodes'' as an incorrect context.

\begin{table}[t]
\centering
\scalebox{0.75}{
\begin{tabular}{llccccc}
\toprule
\multirow{2}{*}{\textbf{Dataset}} & \multirow{2}{*}{\textbf{Type}} & \multicolumn{5}{c}{\textbf{Threshold}}\\
& & 0.5 & 0.6 & 0.7 & 0.8 & 0.9 \\
\hline
\multirow{2}{*}{NQ}
& Answerable & 61.3  &  56.2 & \textbf{34.9} &  6.4 &	0.0\\
& Unanswerable & 2.3 &	27.8 &	\textbf{71.3} &	97.2 &	99.8 \\
\midrule
\multirow{2}{*}{TQA}
& Answerable & 77.3	 & \textbf{51.6}	 & 9.2	 & 0.1	 & 0.0 \\
& Unanswerable & 14.3 &	\textbf{62.7} &	94.7&	100.0&	100.0\\
\bottomrule
\end{tabular}
}
\caption{Performance variation of FiD-KD retriever on answerable and unanswerable sets for different thresholds.
}
\label{tab:appendix_threshold}
\end{table}

\begin{table}
\centering
\scalebox{0.63}{
\begin{tabular}{clcccc}
\toprule
\textbf{P-Generator} & \textbf{R-Generator} &  \textbf{NQ} & \textbf{TQA} \\
\hline
\multirow{3}{*}{\makecell{$\text{Llama2}_{70b}$ \\ (N31.1/T64.3)}}
& $\text{Llama2}_{7b}$    &  36.1 $\rightarrow$ \colorbox{orange}{35.8(-0.3)} & 58.4 $\rightarrow$ \colorbox{lime}{62.8(+4.4)} \\
& $\text{Llama2}_{13b}$   &  38.8 $\rightarrow$ \colorbox{orange}{36.9(-1.9)} & 64.9 $\rightarrow$ \colorbox{lime}{65.4(+0.5)} \\
& $\text{Llama2}_{70b}$   &  40.7 $\rightarrow$ \colorbox{orange}{37.4(-3.3)} & 66.3 $\rightarrow$ \colorbox{orange}{66.2(-0.1)} \\
\midrule
\multirow{2}{*}{\makecell{$\text{Llama3}_{70b}$ \\ (N41.3/T75.1)}}
& $\text{Llama3}_{8b}$    & 38.2 $\rightarrow$ \colorbox{lime}{42.1(+3.9)} & 57.6 $\rightarrow$ \colorbox{lime}{66.9(+9.3)} \\
& $\text{Llama3}_{70b}$   & 46.8 $\rightarrow$ \colorbox{orange}{45.6(-1.2)} & 72.1 $\rightarrow$ \colorbox{lime}{74.0(+1.9)} \\
\midrule
\multirow{2}{*}{\makecell{$\text{ChatGPT}$ \\ (N37.7/T72.0)}}
& \multirow{2}{*}{$\text{ChatGPT}$} &  \multirow{2}{*}{45.9 $\rightarrow$ \colorbox{orange}{43.2(-2.7)}} & \multirow{2}{*}{70.7 $\rightarrow$ \colorbox{lime}{72.1(+1.4)}} \\
& & & \\
\bottomrule
\end{tabular}
}
\caption{
The change in EM score when using the cosine similarity score of the FiD-KD retriever for the confidence score, when utilizing LLM's parameter knowledge for a set of low confidence contexts. The thresholds were set to 0.7 for NQ and 0.6 for TQA, as specified in Table 3.}
\label{tab:appendix_selective_knowledge}
\end{table}

\section{Selectively using parametric knowledge with FiD-KD}
\label{appenidx:D}
Table \ref{tab:appendix_selective_knowledge} shows the change in EM score when applying the mixed decoding strategy, using the cosine similarity score of the FiD-KD retriever as the confidence score. For small parameter generators, the EM score is low when applying the FiD-KD retriever to LLMs, which results in a high gain when utilizing parametric knowledge of large parameter models. However, since the classification performance of the FiD-KD retriever is lower than that of RE, even utilizing parametric knowledge does not significantly outperform the baseline performance of parametric knowledge. Especially for more knowledge-intensive tasks such as NQ, the performance loss is substantial.

\section{FiD-KD retriever's performance in ``unanswerable'' scenarios}
\label{appendix:E}
Table \ref{tab:appendix_threshold} shows the performance of the FiD-KD retriever in unanswerable scenarios according to different threshold values. For the FiD-KD retriever, it is observed that while trying to maintain performance on the answerable set, the classification ability on the unanswerable set significantly decreases.

\end{document}